# RAISE-Vision: An Automated Contact Angle Measurement Platform for Contact Lens Wettability Characterization

Erin Ng[1], Mohammad Nazeri[2], Julie S Liu[1,3], Tithi Choksi[4], Lyndon Jones[5,6], Frank Gu[1,2,7*]

[1]Department of Chemical Engineering and Applied Chemistry, University of Toronto, Toronto, Canada

[2]Institute of Biomedical Engineering (BME), University of Toronto, Toronto, Canada

[3]Department of Chemical Engineering, McMaster University, Hamilton, Canada

[4]Department of Mechanical & Industrial Engineering, University of Toronto, Toronto, Canada

[5]School of Optometry and Vision Science, University of Waterloo, Waterloo, Canada

[6]Centre for Ocular Research and Education (CORE), University of Waterloo, Waterloo, Canada

[7]Acceleration Consortium, University of Toronto, Toronto, Canada

*Corresponding author: f.gu@utoronto.ca

**Abstract:**

Contact angle (CA) is a common parameter measured in the development of novel contact lens materials and surface coating technologies. Existing literature on in vitro CA of contact lenses via the sessile drop method is hindered in repeatability, reliability, and comparability due to inter-laboratory and inter-operator protocol variation. In this work, the Robotic Autonomous Imaging Surface Evaluator for vision applications (RAISE-Vision) is introduced as an automated platform for CA measurement on curved hydrogel surfaces, with specific application to in vitro contact lens wettability analysis throughout air exposure. Customized 3D printed hardware, such as lens mounts, camera holders and drying apparatuses, were designed to facilitate an automated repeated sessile drop workflow within a liquid handler. Collected images were directly fed to a custom open-source image processing script for CA determination on a curved surface that combines the elliptical fit and tangent methods to allow for variations in droplet shape. The CA of senofilcon A contact lenses measured using the RAISE-Vision platform increased from 62.6 ± 5.5° to 83.4 ± 2.1° over 10 sequential measurements in air and was able to replicate results obtained using an analogous manual protocol ($p = 0.470$). The custom image processing script was also separately demonstrated to generate comparable results to a commercially available CA determination software ($p = 0.343$). Aside from minimizing operator variability for CA measurement, RAISE-Vision provides a platform for standardized in vitro contact lens wettability analysis and understanding of surface wetting behaviour during air exposure.

## 1.0 Background/Introduction

Over 150 million people worldwide use contact lenses as their primary form of vision correction or as therapeutic treatments for various diseases [1], [2]. Despite the large investment to advance contact lens technology, 50% of wearers experience contact lens discomfort in varying levels of intensity, which can lead to reduced wear time and ultimately discontinuation of use [3], [4], [5]. Though many bulk properties of soft contact lenses, such as equilibrium water content (EWC), mechanical modulus, coefficient of friction and oxygen permeability, have been examined with respect to discomfort [6], [7], contact lens surface characteristics are also relevant due to their direct interaction with the tear film [5], [8], [9], [10].

Wettability is an important surface property of biomaterials that describes the ability of a liquid to spread over a solid surface [11], [12], [13]. Increasing wettability of contact lenses is generally desirable as it is thought to improve tear film stability and vision quality, while reducing friction with ocular surfaces that may otherwise contribute to sensations of discomfort during wear [3], [6], [14], [15]. Contact lens wettability is of particular interest, given the widespread use of silicone hydrogel (SiHy) contact lenses, which account for 75% of new daily wear soft lens fittings [16]. The siloxane groups that greatly improve the oxygen permeability of SiHy materials are also commonly attached to hydrophobic side chains, which have led to decreased surface wettability outcomes compared to conventional hydrogels [12], [17], [18], [19]. Consequently, advancements in contact lens technology, including the use of internal wetting agents, water gradient technology and plasma surface modification, have been employed to mask hydrophobic groups at the lens surface and improve wettability [10], [17], [20], [21], [22].

The evaluation of contact lens surface wettability can be done via clinical assessments that analyze wettability on the eye or via laboratory-based contact angle (CA) measurements [23]. Non-invasive break-up time (NIBUT), measured via interferometry and videokeratoscopy, is the most common method to assess in vivo contact lens wettability [19], [23], [24], [25]. The quality and stability of the pre-lens tear film can be determined by the time to distortion or breaks in an image projected onto the lens [19], [24], [26], [27]. Measuring in vivo wettability is attractive as the effects of the full physiological environment are captured. However, these studies can have high inter-subject variability, with higher logistical, time and cost burden [19], [25], [26], [28].

The in vitro assessment of wettability is evaluated using CA measurements between a liquid and a solid at the solid-liquid-air interface, with low CA indicating high wettability [29], [30]. For contact lenses, maintaining lower CA over a longer time scale is associated with increased ability of tear spreading over the lens surface, and tear film stability [31]. Young's equation (Eq. (1)) describes the equilibrium CA ($\theta$) as a balance between three interfacial tensions: liquid-solid ($\gamma^{SL}$), liquid-gas ($\gamma^{LV}$), and solid-gas ($\gamma^{SV}$) [13], [32].

$$cos\theta = \frac{\gamma^{SV} - \gamma^{SL}}{\gamma^{LV}} \tag{1}$$

For contact lenses and ocular systems, these interfacial tensions can be altered by lens material [33], contact lens solution [34], blister pack solution formulation [30], [35], biological tear make up [23], [36] and environmental conditions [37], [38]. Although, Young's equation provides the theoretical model for CA measurements, it assumes an ideal solid surface that is flat, smooth, chemically homogenous and nondeformable by the liquid [39], [40]. Since real surfaces, like contact lenses, deviate from these assumptions, experimental determination of CA is often done through geometric or forced based techniques [31], [41], [42]. There is currently no ISO standard for contact angle measurement of soft contact lenses [19], [31], [41], though three main techniques are commonly used, each with their own advantages and disadvantages: Wilhelmy plate, captive bubble and sessile drop [31], [41].

The Wilhelmy plate method is a force-based tensiometry technique used to determine dynamic CA [34], [43]. This technique mounts a flat strip cut from a contact lens to the arm of a high-precision balance over a liquid bath [34]. Instead of relying on visual metrics, the micro-balance measures the dynamic force change, as the sample is moved in and out of the fluid to calculate advancing and receding CA [34], [41]. A few drawbacks of this method include, the intricate and destructive sample preparation, which can introduce contamination, dehydration, stretching or tearing, and the time-consuming experimental method compared to the sessile drop and captive bubble techniques [18], [41], [44].

Captive bubble is another established in vitro CA measurement technique for hydrated contact lens samples [18], [41], [45]. In this configuration, the contact lens is fully immersed in an aqueous medium, such as distilled water or a simulated tear fluid [18], [45], [46]. An air bubble is subsequently dispensed from a curved syringe beneath the sample and allowed to adhere to the downward facing anterior lens surface [18]. Because the lenses remain fully hydrated during the measurement, this method provides smaller measurement errors compared to the sessile drop method, where lenses are susceptible to dehydration [46]. However, lenses can maintain an absorbed layer of liquid on the surface that the air bubble does not replace, leading to poor differentiation between contact lens materials [41].

Despite the advantages of clinical and submerged testing, the sessile drop method remains the most widely used technique for in vitro wettability measurement, due to its simple setup and equipment requirements [19], [41], [45], [47]. The technique involves removing a contact lens from its storage medium, carefully blotting it to remove excess liquid, and mounting it on the stage of a goniometer [19], [48], [49]. After centering the lens directly under a computer-controlled syringe, a droplet of liquid is placed on the lens surface, and the CA between the solid and the liquid can be determined with imaging and computer software [50], [51]. Some programs can determine CA automatically, while others require user input to define the droplet and baseline profiles, which could lead to difficulty in reproducibility and poor accuracy [31], [51], [52]. Overall, many of the commercial image processing software that are supplied with the goniometer instruments are capable of calculating CA via various methods, including circle fitting, ellipse fitting, tangent method, Young-Laplace fitting and half angle ($\theta/2$) method [39], [46], [53]. Read et al. have also previously developed a customized MATLAB-based image analysis routine to automatically determine the advancing and receding CA for contact lenses captive bubble configuration, demonstrating the potential for tailored CA determination approaches [54].

CAs determined via the sessile drop method are often higher than those measured via the captive bubble or Wilhelmy plate method, as the measurement is conducted in air exposure [35]. However, the main reproducibility challenge of the sessile drop method lies in inter-laboratory and inter-operator variations in sample preparation and protocol execution [31], [41], [45], [46]. Figure 1 depicts a categorical comparison of detailed sessile drop method protocols for the assessment of CA of soft contact lenses reported in the literature, highlighting differences in drying method, time to measurement, measurement temperature, droplet volume, and surface shape.

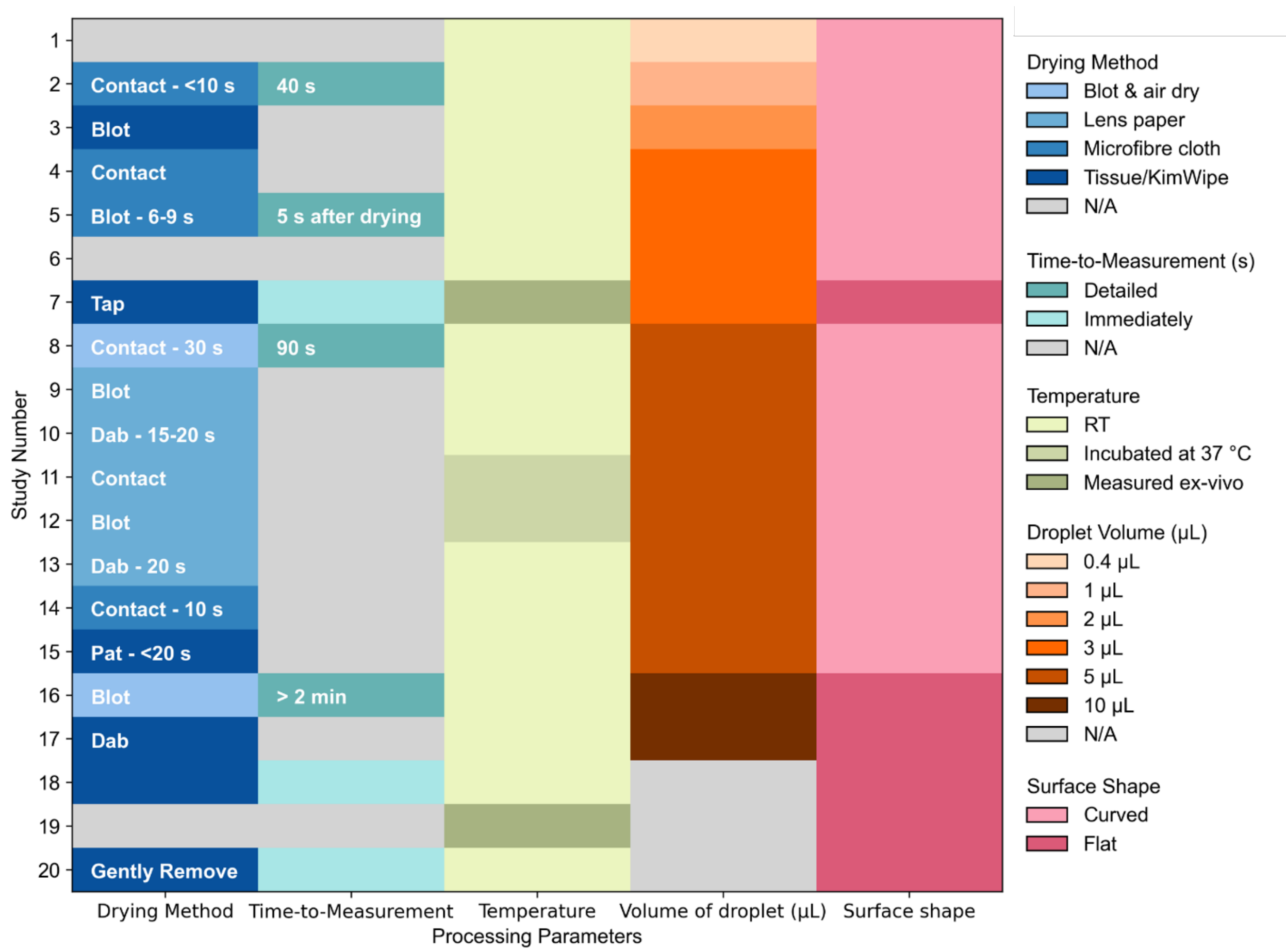

Figure 1. Categorical comparison of sessile drop method protocols for contact lenses reported in the literature. Details provided about the drying method and time to measurement are overlayed as text [12], [18], [19], [33], [35], [46], [47], [50], [51], [55], [56], [57], [58], [59], [60], [61], [62], [63], [64], [65].

As previously mentioned, the initial drying protocol, which is performed to remove excess liquid from the lens surface before CA measurement, varies widely [41]. Aside from using a variety of materials for absorbing liquid, such as lens paper, KimWipes, and microfibre cloths, many of the drying actions are often described with subjective terminology like ‘pat’, ‘tap’, ‘dab’, or ‘blot’, which can be highly operator dependent [18], [35], [56], [57], [58]. Some studies specify the number of repeated actions or the duration of the drying protocol to minimize variation in protocol execution [46], [47], [50], [51], while others omit details about sample preparation all together [33], [59], [60]. Time to measurement is also significant for a sessile drop CA measurement, as the hydrogel surface chemistry rearranges in consequence to air exposure and surface dehydration [46], [51], [61]. However, inconsistent reporting practices and variations in this interval persist in the field, which may limit the repeatability and comparability of data reported in the literature. For example, Rickert et al. allow the contact lens material to air dry for 2 minutes after blotting, while Read et al. allow 5 seconds after blotting before CA measurement [46], [62].

Within the reviewed protocols, droplet volume typically ranged between 3-5 µL, with a few studies using smaller volumes, which are more susceptible to disturbances during dispensing [66]. The sessile drop method was also mainly carried out on the native curved anterior surfaces of contact lenses, which was not surprising, as flat CA measurement would either require destructive alteration to the lens [55], [58], [60], [63], or for a separate flat sample to be prepared [62]. In the case of CA measurement on curved surfaces, it is critical that the image analysis accounts for the curved lens baseline, to ensure the actual CA, intrinsic to the material, is being reported rather than the apparent CA relative to a horizontal baseline [67], [68].

This lack of harmonization in sample preparation and measurement method protocol makes it difficult to draw direct comparisons about contact lens material performance across the literature [33], [41], [46]. For example, Giraldez et al., Chwalik-Pilszyk et al., and Phan et al. (mapped to study 20, 1 and 10 in Figure 1), have measured the water CA of senofilcon A directly from its blister pack solution via their implementation of the sessile drop method to be 55.15 ± 6.87°, 76.3 ± 8.1°, and 96.4 ± 2.5° respectively [33], [63], [64]. Their methodologies varied in drying method, droplet volume and surface geometry. Chwalik-Pilszyk et al. used 0.4 µL of distilled water compared to 5 µL used by Phan et al., and Giraldez et al. tested on a flat contact lens, which provided varied results regarding the hydrophobicity of the lens material.

As hydrogel materials are sensitive to variabilities during sample preparation and changes in its local environment [46], the majority of disadvantages of the sessile drop method stem from varied technique and operator interpretation of protocols, highlighting the need for standard procedures [41]. A system that alleviates some operator dependence and variability would be beneficial to advancing the understanding of in vitro contact lens wettability.

Surface characteristics of SiHy contact lens materials over air exposure are of interest as accumulation of desiccation stress can induce changes to the structure of employed wetting technologies or migration of mobile hydrophobic moieties to the surface, affecting wetting properties [10], [61]. Eftimov et al. measured contact lens wettability profiles over 16 hours of continuous blink-like air exposure (10 s) and rehydration (1 s) cycles, to simulate the drying stress of daily wear [61]. This study determined that CA increased after air exposure for all materials: 68.6° to 94.5° for senofilcon A, 32.1° to 89° for stenfilcon, and 23.5° to 49.3° for delefilcon A [61]. Ketelson et al. observed comparable trends when measuring water CA of conventional hydrogel lenses as a function of varied wetting and drying cycles (5 min soak/1.5 min air exposure, 30 min soak/1.5 min air exposure) over 1-5 hours [51].

Though some of these studies used automated models to condition the lenses over an extended period of time, the sessile drop method protocol remains largely manual, requiring continuous hours of operator involvement for sample handling, preparation, measurement on a benchtop goniometer and image analysis. These scalability and consistency issues can be addressed by laboratory automation, which is well suited to repetitive tasks, allowing operators to focus on more intellectually demanding work [69], [70]. Automated systems would allow for standardized drying protocols and controlled timing between measurements for streamlined analysis of contact lens wettability over time [71]. Most goniometer systems offer automated syringe control and semi-automated image analysis [46], [51], [56], but a gap still exists in automated sample handling and preparation. With CA results of in vitro contact lens wettability analysis being methodologically dependent [18], [25], [30], and only limited reports of fully automated CA measurement techniques, there exists an opportunity to improve the understanding of contact lens wettability and early-stage development of novel contact lens materials via laboratory automation.

In this work, we report an automated sessile drop CA measurement platform, RAISE (Robotic Autonomous Imaging Surface Evaluator)-Vision, to improve the standardization and to reduce operator dependence of contact lens wettability analysis. Among the three main in vitro contact lens wettability measurement techniques, sessile drop was selected as the focus, due to its wide use throughout the field, non-destructive sample handling, and protocol compatibility with standard liquid-handler deck space and capabilities. The present study builds on the RAISE platform previously reported by Nazeri et al., which demonstrated CA measurement automated within a liquid handler robot as part of a self-driving lab linking wettability to liquid formulation design in a closed-loop [72]. The RAISE-Vision platform tailors this wettability measurement approach for contact lenses by integrating custom tools to achieve consistent sample handling (blotting and droplet removal for sequential CA measurement) and automated image analysis on curved surfaces without any operator input. Within this study, the in vitro wetting performance of senofilcon A was evaluated at room temperature over air exposure using the automated RAISE-Vision platform.

## 2.0 Materials & Methods

### 2.1 Contact Lenses

Commercial daily disposable silicone hydrogel contact lenses (senofilcon A material) were used as the benchmarking material in this study. The power, base curve and diameter of the lenses were -1.0, 8.5 mm and 14.0 mm respectively. The water content and oxygen permeability listed by the manufacturer, Johnson & Johnson (Jacksonville, Florida), was 38% and 121 Dk/t [7], [73], [74].

### 2.2 Custom 3D Printed Parts

The custom hardware for the stage, camera holder, lens mount, and DryPads were designed using a computer-aided design (CAD) program. The designs were sliced to a layer height of 0.2 mm and 3D printed via fused deposition modeling (FDM) on a Bambu Lab X1E (Bambu Lab, Shenzhen,

China) equipped with a 0.4 mm nozzle. Parts were printed with polylactic acid (PLA) filament at an infill density of 15%.

## 2.3 Automated Wettability Analysis

The RAISE-Vision platform is presented as a means of standardizing wettability analysis across the contact lens field and involves both hardware and software developments to physically carry out the experiment and to automate analysis. In this version of RAISE-Vision, static CA was measured using a single dispensed droplet, consistent with other sessile drop CA measurements in contact lens studies [12], [64]. To capture changes in wettability over time, static CA measurements were repeated during air exposure, which provided a temporal profile of the CA. All files associated with the RAISE-Vision platform have been uploaded to GitHub (https://github.com/Frank-Gu-Lab/RAISE-Vision).

### 2.3.1 Hardware Development

The sessile drop CA measurement was automated within an Opentrons OT-2 liquid handler (Opentrons Long Island City, USA) fitted with a 20 µL pipette to examine contact lens wettability over air exposure. The workflow relies on a suite of customized hardware, including the stage, lens mount, and drying adapters, to support imaging, substrate handling and to perform drying protocols, shown in Figure 2.

*Translation from RAISE to RAISE-Vision Stage*: The initial stage for RAISE-Vision featured a staircase design adapted from the RAISE platform, developed by Nazeri et al., originally intended for high-throughput CA measurement on flat substrates [72]. Six CA measurement locations were specified on the two levels on the RAISE-Vision stage (Figure 2a), which were designated by housing slots for custom convex lens mounts. With a base curvature of 8.5 mm, diameter of 14.0 mm and height of 34.6 mm, the lens mount aimed to eliminate the air gap between the lens and the mount and served as a lens handling tool that easily fit in the blister packaging of contact lenses and a standard 24-well plate. This stage configuration necessitated a mobile camera to image the lens and applied droplets at varied locations and levels. The press-fit camera holder design introduced by Nazeri et al. enabled this capability [72]. The 3D printed module interfaced with the right-pipette on the Opentrons OT-2 gantry, which allowed the liquid handler to control the position of a pen-like camera (12 mm diameter, Opti-Tekscope, Phoenix, USA) via native pipette movement functions. Three M4 screws and nuts at the bottom of the holder secured and prevented misalignment of the camera as it moved over the deck space. The light sources for CA images were provided from the front via a ring light on the camera, and from the back via a coiled LED strip secured in the staircase stage design.

*Optimized RAISE-Vision Stage:* Following evaluation of the initial platform design, modifications were implemented to address observed challenges (described in Section 4.1). The succeeding version of the RAISE-Vision platform opted for a condensed imaging stage that sits within a single deck slot of the liquid handler (Figure 2b). This stage houses a stationary camera mounted via push fit and supports a single seat for a lens mount, which was elevated to ensure the apex of the lens was imaged in profile. An LED panel (RaLeno, Shenzhen, China) was positioned behind the lens mount to enhance image contrast. Though the base curvature and diameter of the lens mount remained the same as the previous iteration, the height of the lens mount was decreased to 13.8 mm. The assembly of this component was adjusted into a plate geometry, with the shortened lens mount secured by a cylindrical transition fit onto an extended handle to aid substrate handling. Further, instead of 3D printing with PLA, the lens mount was machined of aluminum, providing a smooth surface and closer contact to the lens materials.

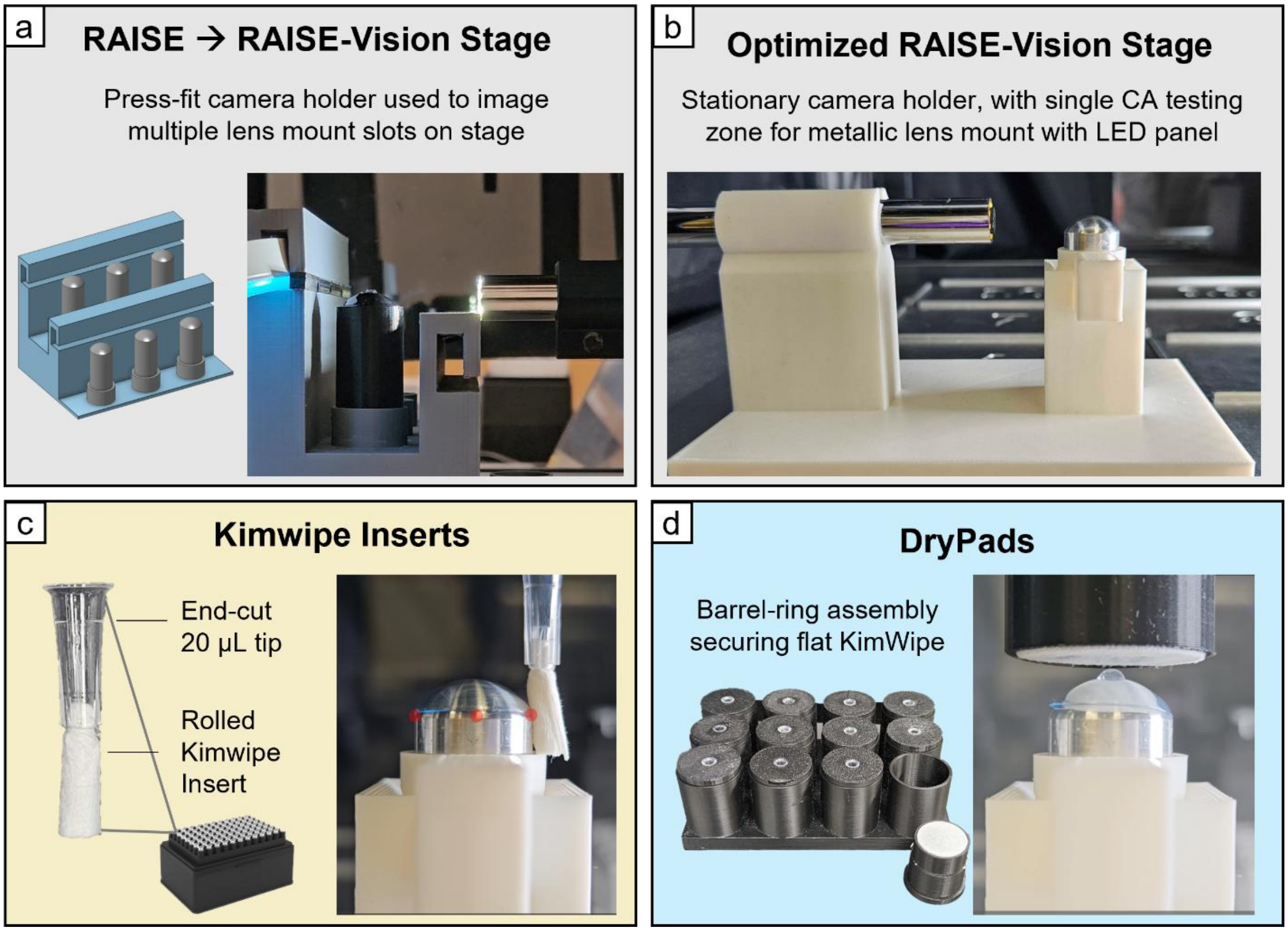

Figure 2. a) Initial iteration of RAISE-Vision stage featuring six CA measurement locations over two levels, and LED strip backlighting, which required a mobile camera for imaging, b) Optimized iteration of RAISE-Vision stage featuring a single lens mount, with a stationary camera that used a large LED panel for high contrast lighting, c) Breakdown of custom KimWipe Insert hardware, featuring a rolled Kimwipe insert inside an end-cut 20 µL pipette tip. The KimWipe insert is used within the initial drying protocol to wick excess liquid from the lens at four equally spaced points of the lens mount (3/4 are shown in red), d) Close up of custom pick-and-place DryPad hardware, used for removing water droplets from the contact lens surface between CA measurements.

*Drying Apparatuses:* These designs build on the pick-and-place tool change strategy developed by Nazeri et al. [75], which allows custom modules to interface with standard pipettes, adding functionality to the liquid handling system. These custom tools can easily be manipulated by the pipette gantry using the robot's native API functions.

The Kimwipe inserts (Figure 2c) consist of a tight roll of Kimwipe, 2 mm in diameter inserted into an end-cut 20 µL pipette tip and cut to 10 mm length. This apparatus is used in the initial drying protocol to absorb excess liquid from the contact lens prior to CA measurement. These custom components were arranged within a standard 20 μL tip rack, facilitating simple integration to the liquid handler without defining a new piece of labware.

The DryPads (Figure 2d) are a 3D printed custom tip assembled from a 20.6 mm diameter barrel and close running fit ring (21.0 mm inner diameter) sandwiching two layers of 1-ply Kimwipe cut to approximately 3cm x 3cm. This hardware was used to remove droplets from the lens surface during the sessile drop workflow, allowing repeated CA measurements throughout air exposure to be made. An end-cut 20 μL pipette tip is mounted inside the barrel adaptor, enabling compatibility with the liquid handler. A holder, designed to fit within a single deck slot on the OT-2, accommodates 12 DryPads at a time.

*Custom JSON Files:* To allow the liquid handler to manipulate and act on the custom hardware, JSON files defining the stage, DryPads and trash location designated for KimWipe inserts (*raise_vision_stage.json*, *raise_vision_drypad_tiprack.json, raise_vision_trash.json*) were generated using Opentron's Custom Labware Creator software. These files describe the labware dimensions, labware type (well plate for stage, tip rack for DryPads, tip rack for trash), accessible well locations, x-y-z offsets and tip lengths (applicable to DryPads only) of the hardware not already registered as a standard labware in the Opentrons' Labware Library.

### 2.3.2 Integration of Camera

To automate image capturing at specific points of the CA measurement workflow, the Opentrons liquid handler and USB camera were integrated via an orchestrator server run by a computer acting as the main control console. By connecting the Opentrons OT-2 and the computer to the same local area network (LAN), the computer acts as a server that runs an orchestrator program (*CameraServer.py*) which opens a socket and establishes real-time communication with both the liquid handler and camera. The camera operates as an auxiliary module of the OT-2, enabling image capturing functions to be called directly within the robot's protocol script. This integration ensures the liquid handler and camera can be controlled concurrently in a synchronized and automated workflow.

### 2.3.3 Workflow

The actions carried out by the liquid handler during the contact lens wettability analysis were specified by a script developed using the Opentrons Python API uploaded into the OT-2's onboard Jupyter Notebook environment (*RAISE_Vision.ipynb*). Custom labware JSON files were also uploaded into this onboard environment such that the OT-2 recognized and used these designs in the executed protocol. A block flow diagram detailing the actions performed at every stage of the automated RAISE-Vision workflow is presented in Figure 3a. The workflow detailed here is centered around the deck layout represented in Figure 3b, including the updated RAISE-Vision stage, featuring a stationary camera, well plate, tip racks, DryPads and trash slot. A similar protocol was previously carried out for earlier iterations of the RAISE-Vision platform.

To start, the lenses were manually removed from their blister pack solution by inverting and dipping the lens mount into the packaging to pick up the lens. Once the lens mount was placed into its seat on the stage, the automated workflow began.

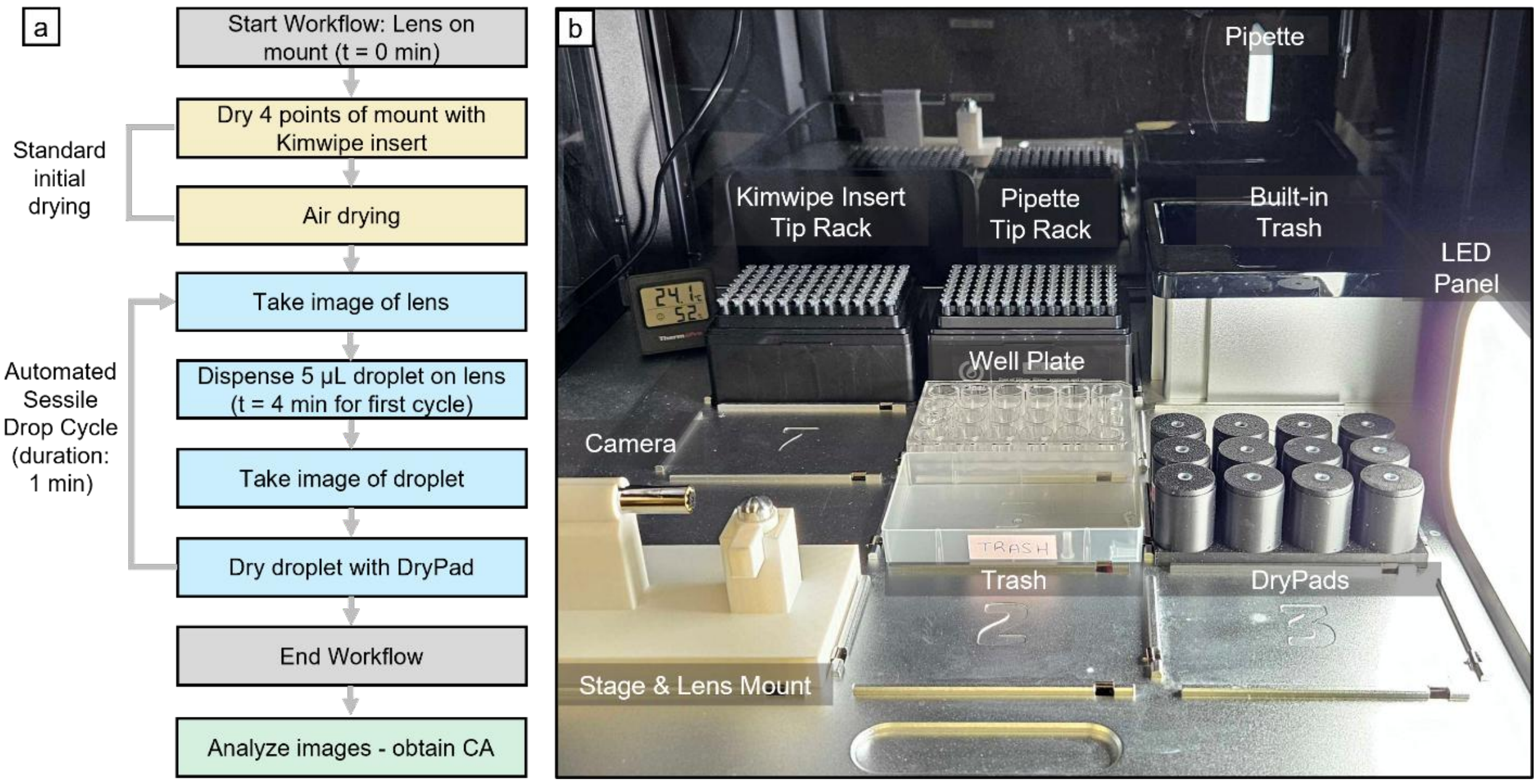

Figure 3. a) Block flow diagram of the actions carried out during the automated sessile drop CA measurement assay (RAISE-Vision), b) Deck set-up for RAISE-Vision platform

The initial drying protocol, referred to as *drySides()* in the RAISE-Vision workflow script, was developed as a standardized method to remove excess liquid from the lens, without altering the surface at the apex of the lens. With a Kimwipe insert attached to the pipette, the gantry contacted the lens mount at four equally spaced points, for 5 seconds each, to absorb most of the excess fluid on the lens. The used Kimwipe insert was deposited into a designated trash slot rather than the liquid handler's built-in trash bin to allow for simple replacement of the custom tip and to prevent contamination with other tips. During the *drySides()* action, the lens was exposed to air over 50 seconds. Additional moisture on the surface of the lens was passively removed by allowing it to air dry for a total of 4 minutes at room temperature before the first droplet was applied.

After the initial drying protocol, the automated sessile drop workflow ensued. First, an image of the lens before dispensing the droplet was taken for image processing purposes. Then, the liquid handler's 20 µL pipette formed a 5 µL droplet of Milli-Q water with a flowrate of 1 µL/s on the pipette tip over the center of the lens. The gantry moved downwards at 2 mm/s until the droplet was transferred onto the contact lens. The pipette was lifted and after settling for 5 seconds, a second image of the lens and the dispensed droplet was taken. The pipette tip was disposed of into the built-in trash bin. Then, a fresh DryPad apparatus was picked up with the 20 µL pipette and centered over the lens mount. The gantry moved downwards to gently contact the lens surface, removing the applied droplet and preparing the substrate for a subsequent CA measurement. After the DryPad was returned to its holder, the sessile drop cycle was repeated. Each CA measurement was made 1 minute apart, allowing for consistent temporal wettability analysis.

This platform allows wettability to be measured as a dynamic property rather than as a single CA value. In this study, RAISE-Vision was used to evaluate the wetting profile of senofilcon A, a SiHy contact lens material, over 10 sessile drop cycles at room temperature. Lenses were tested in triplicate.

### 2.3.4 Automated Image Analysis

Images collected during the sessile drop CA workflow were directly processed using a custom Python-based image analysis script for curved surface CA analysis without operator input. The workflow can be divided into two stages: general image processing and CA determination, which are visually presented in Figure 4. The program uses Python's scikit-image, SciPy, NumPy and Matplotlib libraries for image processing, boundary tracing, curve fitting, CA determination and visualization.

Two images were taken during each sessile drop CA measurement: an image of the lens prior to dispensing a drop (pre-lens image) and an image of the droplet on the contact lens surface (droplet image). The following processing steps were applied to all images before analysis. First, the images were cropped to isolate the region of interest. Then, the images underwent grayscale and gaussian blurring filtering ($\sigma = 5$) to simplify the image and remove granular artifacts at contour edges. A gamma correction ($\gamma = 5$) was used to address image distortion caused by lens texture and reflection [76]. Lastly, Otsu thresholding was applied to generate a binary image that separates the lens (and droplet) in the foreground from the background.

To accurately determine the CA of the droplet, the lens baseline curve must be accounted for [67]. Using scikit-image's measure module, the main contour of the pre-lens image was found and stored as a 2D array. The lens baseline curve was approximated to an ellipse using SciPy's optimize module on the pre-lens image, and the fitted parameters and root mean square error (RMSE) were saved to a text file. For the droplet image, the main contour represents both the lens and droplet. As such, the fitted lens baseline curve was superimposed on the droplet image, after which the droplet contour was isolated. A mask was applied to identify when the droplet image's main contour deviates appreciably from the lens baseline curve, characterized by vertical displacement ($\Delta > 5$ pixels) and slope disparity ($\Delta > 1.5$). A minimum segment length of 70 pixels was set to ensure that artifacts on the lens surfaces were not detected. The resulting array defined the droplet contour. Two methods of determining CA commonly used by commercial software were employed: the elliptical fit method and tangent method.

*Elliptical Fit Method:* Analogous to the lens baseline curve, the isolated droplet contour was fit to an ellipse using SciPy's optimize module. The coordinates of the left and right solid-liquid-air interfaces were identified by numerically solving the coupled implicit equations of the droplet and lens ellipses. Normal vectors of the lens curve ($l$) and droplet ellipse ($d$) at the points of intersections were determined and using the dot product, the left and right CA were calculated via Eq. (2):

$$cos(\theta) = \frac{l \cdot d}{|l||d|} \tag{2}$$

The tangent lines, perpendicular to the corresponding normal vectors, were visualized over top the droplet ellipse and lens baseline curve.

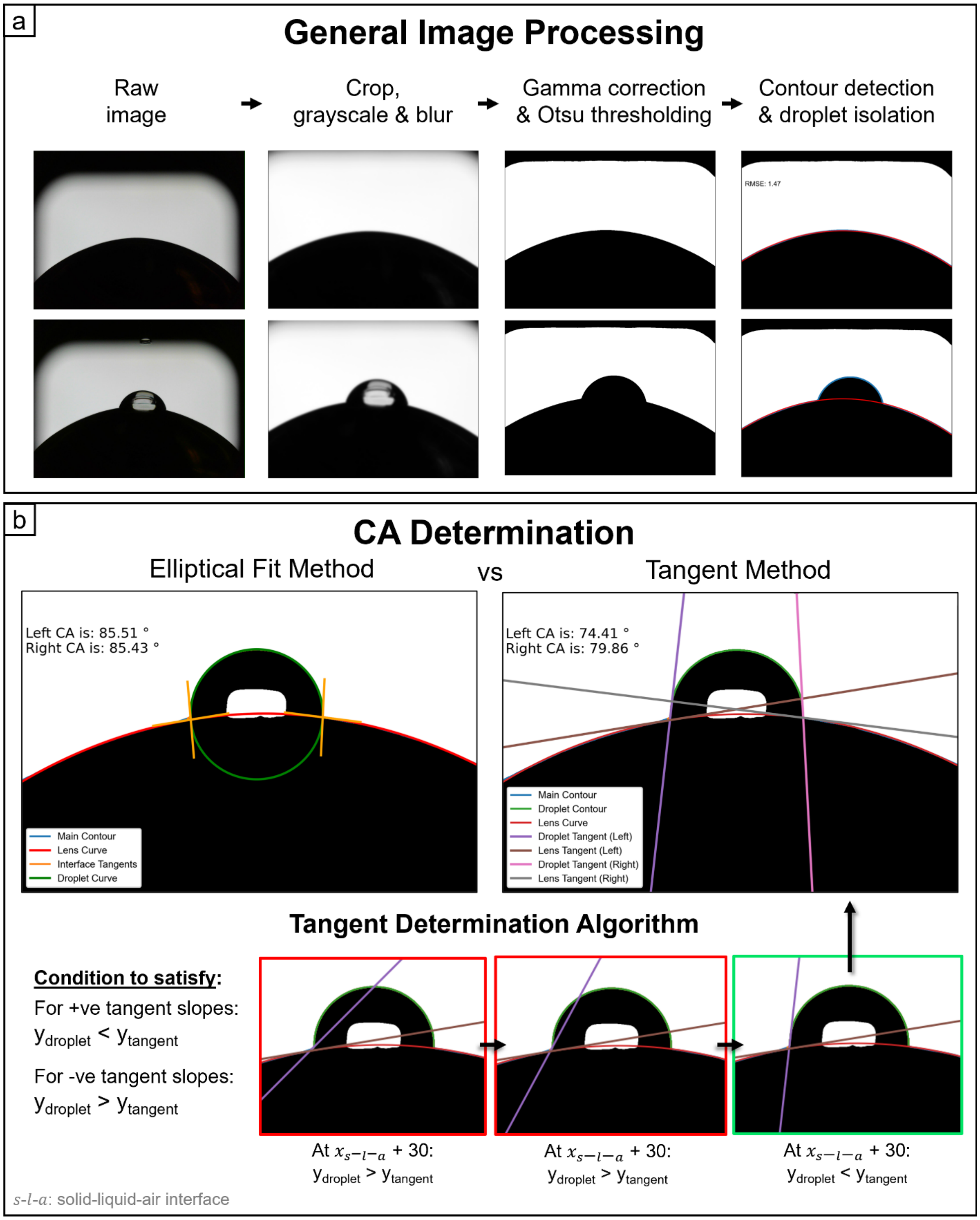

Figure 4. Automated image analysis for CA measurement on curved surfaces broken down into a) general image processing steps, which include image cropping, grayscale and gaussian filters, gamma correction, thresholding, contour detection and droplet separation; and (b) CA determination over a curved surface via an elliptical fit method and tangent method. The algorithm used to verify the accuracy of the generated droplet vector at the solid-liquid-air interface is also described.

*Tangent Method:* For the tangent method, the two endpoints of the identified droplet contour were considered as the left and right solid-liquid-air interfaces. Tangent vectors at both intersections were found for the lens curve $(l)$ and the droplet contour $(d)$. Left and right CAs were calculated using Eq. (2).

Due to methodological limitations of the droplet contour isolation, the identified endpoint was not always accurate, sometimes allowing the tangent vector, $d$, to cross over the droplet region. An automated algorithm was developed to verify the accuracy of $d$ by comparing the y-values of the droplet contour and calculated tangent line 30 pixels from the solid-liquid-air interface. Based on the sign of the tangent slope, a different condition was assessed to determine if $d$ crossed over the droplet region. For the left side, if the droplet's tangent slope was positive, $d$ was identified as valid when the y-value of the tangent was greater than the y-value of the droplet contour (Figure 4). If the tangent slope was negative, $d$ was identified as valid when the y-value of the tangent was less than the y-value of the droplet contour. For the right side, the opposite conditions were established.

In situations where $d$ did cross over the droplet region, the droplet contour was cut by 1 pixel on the corresponding side (left or right) to create two new tangent vectors, $l$ and $d$, until the specified condition was satisfied. These tangent lines are overlaid on the processed droplet image plot within the script as a qualitative check of the CA analysis accuracy.

For rare situations where the isolated sessile droplet was highly anomalous, the Python script allowed for the manual adjustment of the number of pixels being cut for one or more specified droplets. Tangent lines are also visualized over top the droplet image, such that the manual adjustment may be repeated as needed until a desirable tangent is produced.

The final custom image processing script employed in RAISE-Vision involves the integration of both CA determination methods, though each method can be toggled to run independently. In the combined script, the ellipse method is employed first. The normalized root-mean square error (NRMSE) of the ellipse fit near the contact points was evaluated using the lower 25% of the droplet contour. In the case where NRMSE was greater than 0.005, indicating a poor representation of the droplet at the solid-liquid-air interface, the tangent method was then employed for a more accurate determination of the CA.

Once all images were processed, the left CA, right CA, and lens curve parameters were saved to a user-specified folder, along with all processed images, iterations, and corresponding processing conditions. For each droplet, the reported CA was calculated as the mean between the left and right CA.

## 2.4 Manual Water Contact Angle Measurement

Manual sessile drop water CA measurements were made according to previously published methods [18], [35], [51]. Using tweezers, being careful to only touch the edge of the lens, the contact lens was removed from the blister pack solution and placed anterior side down on two layers of KimWipe. The contact lens was manipulated to facilitate even drying across the lens surface and was turned onto the posterior side to allow excess liquid to drain. The total drying time took less than 20 seconds. The contact lens was placed posterior side down on a curved mount which was then centered directly under the syringe of the Optical Contact Analyzer (OCA 20, DataPhysics Instruments GmbH, Filderstadt, Germany).

One minute after initially removing the contact lens from the blister pack solution, a 5 μL drop of Milli-Q water was dispensed at 2 μL/s onto the lens surface using the computer-controlled syringe. The two visible angles formed between the droplet and a user defined curved baseline were measured and averaged (SCA 20, Version 4.3.19, DataPhysics Instruments GmBH, Filderstadt, Germany).

Senofilcon A lenses were used as a benchmark material to compare the automated and manual sessile drop workflows. CAs were measured every minute over 10 minutes of air exposure at room temperature. Manual measurements were made in triplicate.

Aside from using the commercially available image processing software, images collected from the manual CA measurement were also passed through the custom image processing scripts (combined, elliptical fit and tangent) to assess the efficacy of the automated image analysis.

### 2.5 Statistical Analysis

Statistical analysis was conducted using R's afex, and emmeans libraries. A two-way repeated measures analysis of variance (ANOVA) was conducted for the benchmarking CA data with testing method (manual vs RAISE-Vision) as a between-group factor, and cycle number as a within-group factor. A similar two-way repeated measures ANOVA was conducted with image processing method as a between-group factor, and cycle number as a within-group factor for the manually collected droplet images and RAISE-Vision droplet images. Statistical significance was considered when $p < 0.05$.

## 3.0 Results

### 3.1 Benchmarking RAISE-Vision Assay

The wetting profile of the senofilcon A lenses were measured directly from their blister pack solution at room temperature using the RAISE-Vision platform and the described manual workflow. The experiment was conducted in triplicate at a relative humidity of 42%. Images of the experimental setup and the wetting profiles observed using the manual protocol and automated RAISE-Vision platform are shown in Figure *5*. The ANOVA summary comparing the effect of cycle number and testing method is reported in Table 1. Each sequential CA measurement was taken 60 seconds after the previous measurement, allowing the lens to dry incrementally throughout the experiment. The initial contact angle was measured as 67.6 ± 11.3° and 62.6 ± 5.5° manually and with RAISE-Vision, respectively. Regardless of workflow, the water CA increased over cycle number ($p < 0.001$), indicating a decrease in contact lens wettability with exposure to air. The CA of senofilcon A lenses increased to 89.2 ± 5.9° and 83.4 ± 2.1° after 10 measurements using the manual workflow and RAISE-Vision, meaning the observed wetting profiles were within the range of reported CA values for senofilcon A directly from their blister pack solution reported in the literature (55.2 ± 6.9° to 95.4 ± 2.5°) [63], [64]. Overall, the effect of testing method on observed wettability profile was not statistically significant ($p = 0.470$). The standard deviation in observed CA at each cycle between the three automated runs generally ranged between 1.2 – 5.5°, which aligns with precision levels achieved by the manual measurement (1.8 – 13.7°) and prior studies [12], [33], [64].

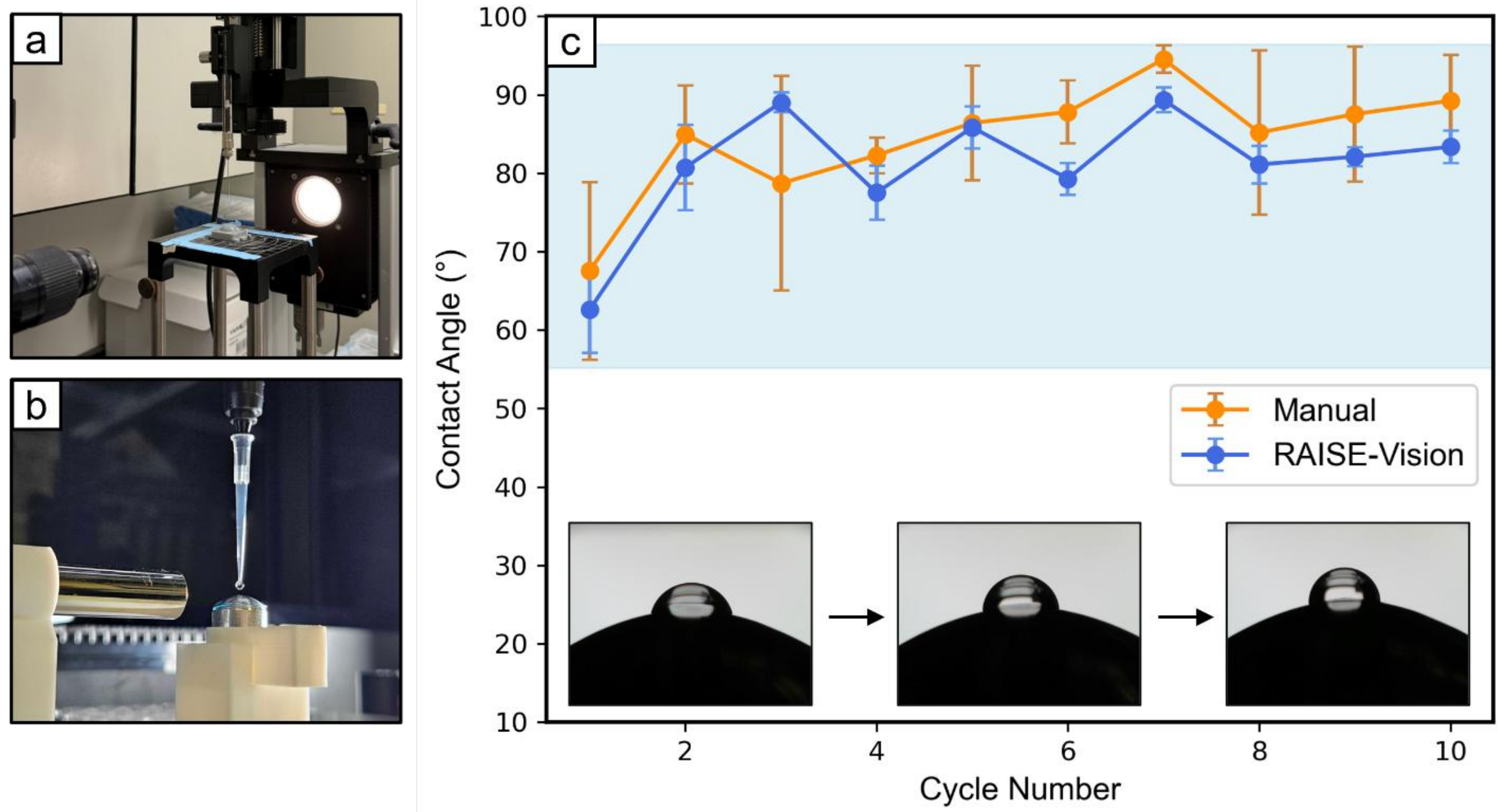

Figure 5. Comparison between manual and automated CA measurements: a) image of goniometer equipment, b) image of CA stage within RAISE-Vision platform and c) wettability profile achieved using manual measurements and RAISE-Vision. The CA range reported in literature for senofilcon A (55.2 -95.4°) is represented by the blue shaded region. Raw images collected over the RAISE-Vision assay are also displayed.

Table 1. Two-way repeated measures ANOVA summary table for benchmarking data.

| Effect | Degrees of freedom | F-statistic | Effect size ($\eta^2 G$) | p-value |
|---|---|---|---|---|
| Testing Method | 1, 4 | 0.64 | 0.033 | 0.470 |
| Cycle | 9, 36 | 5.33 | 0.512 | <0.001 |
| Testing Method × Cycle | 9, 36 | 0.79 | 0.134 | 0.628 |

## 3.2 Comparing Image Processing Methods

To investigate the difference between the elliptical fit and tangent method used within the combined image processing script for determining CA, all images collected from the manual and RAISE-Vision protocols were processed with each method independently. The wetting profiles obtained for each condition are shown in Figure *6*. CA results obtained from the manual testing method did not differ significantly between each image analysis method ($p > 0.05$). It was unsurprising that the commercial image processing software and custom elliptical fit image processing script rendered similar results, as the ellipse-fitting evaluation mode was selected on the SCA20 program to determine CA [77]. In general, the CAs determined with the tangent method were around 5-10° lower on average than the CAs determined using the elliptical fit method, as seen in Figure *6*ab.

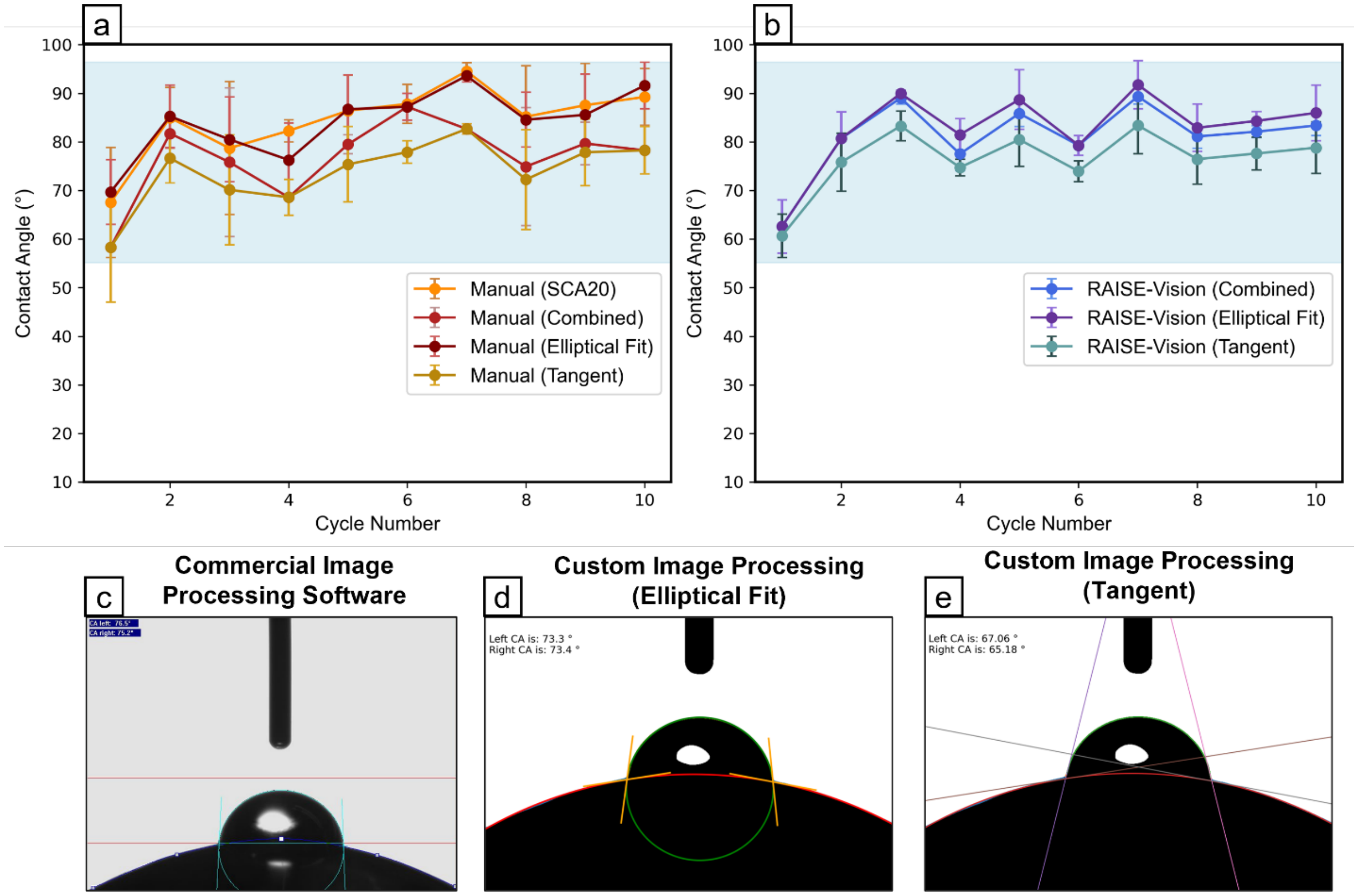

Figure 6. a) Wetting profile obtained using the manual testing protocol and analyzed with the commercial image processing software, and custom image processing scripts (combined, elliptical fit and tangent methods), b) Wetting profile obtained using RAISE-Vision testing protocol and analyzed with custom image processing scripts (combined, elliptical fit and tangent methods), c-e) Images of the same droplet obtained using the manual protocol processed with the commercial image processing software and custom image processing scripts (elliptical fit and tangent method).

## 4.0 Discussion

### 4.1 Iterative Development of RAISE-Vision

When translating the RAISE stage for RAISE-Vision, many of the original features such as the high throughput design, with multiple CA testing locations, original lighting configuration and accompanying mobile camera were retained. However, workflow testing identified opportunities to simplify the design.

First, the press-fit camera holder was not well-suited to the contact lens wettability assay, as movement of the gantry after dispensing the droplet to reposition the camera introduced mechanical disturbance to the sample. This light shaking could cause movement of the original droplet contact line by overcoming pinning forces associated with surface roughness or chemical heterogeneity on the non-ideal contact lens surface [78]. As non-ideal surfaces possess many metastable states for droplet-solid systems, the CA of the disturbed droplet may be different than the droplet 'as-placed', contributing to variability in measurements [40], [78], [79]. Therefore, it was important to minimize disturbances to the sample between droplet dispensing and imaging, prompting the design change towards the stationary camera. In this configuration, only one lens mount was accommodated on the stage. Though the number of lens mount slots was reduced, the effect on experimental throughput was negligible because the RAISE-Vision platform, regardless of stage capacity, can only monitor the CA evolution of one lens at a time. The platform can run for 15 minutes of continuous testing with operator handling required at the start and end of the assay.

In addition to simplifying the stage, the lens mount was redesigned and machined of aluminum to provide consistent contact between the lens and mount, which was not possible with 3D printed parts, as layer-by-layer fabrication created crevices that trapped residual liquid. Because the aluminum mount was shortened to 13.8 mm, a plate-like support was 3D printed to improve the manual handling of the new mount, shown in Figure 2b. This lens mount still easily fits into blister packaging and round 24-well plates, facilitating easy integration of standard labware for future automated testing of different contact lens materials in soaking and incubation studies common to the field [12], [62], [80], [81].

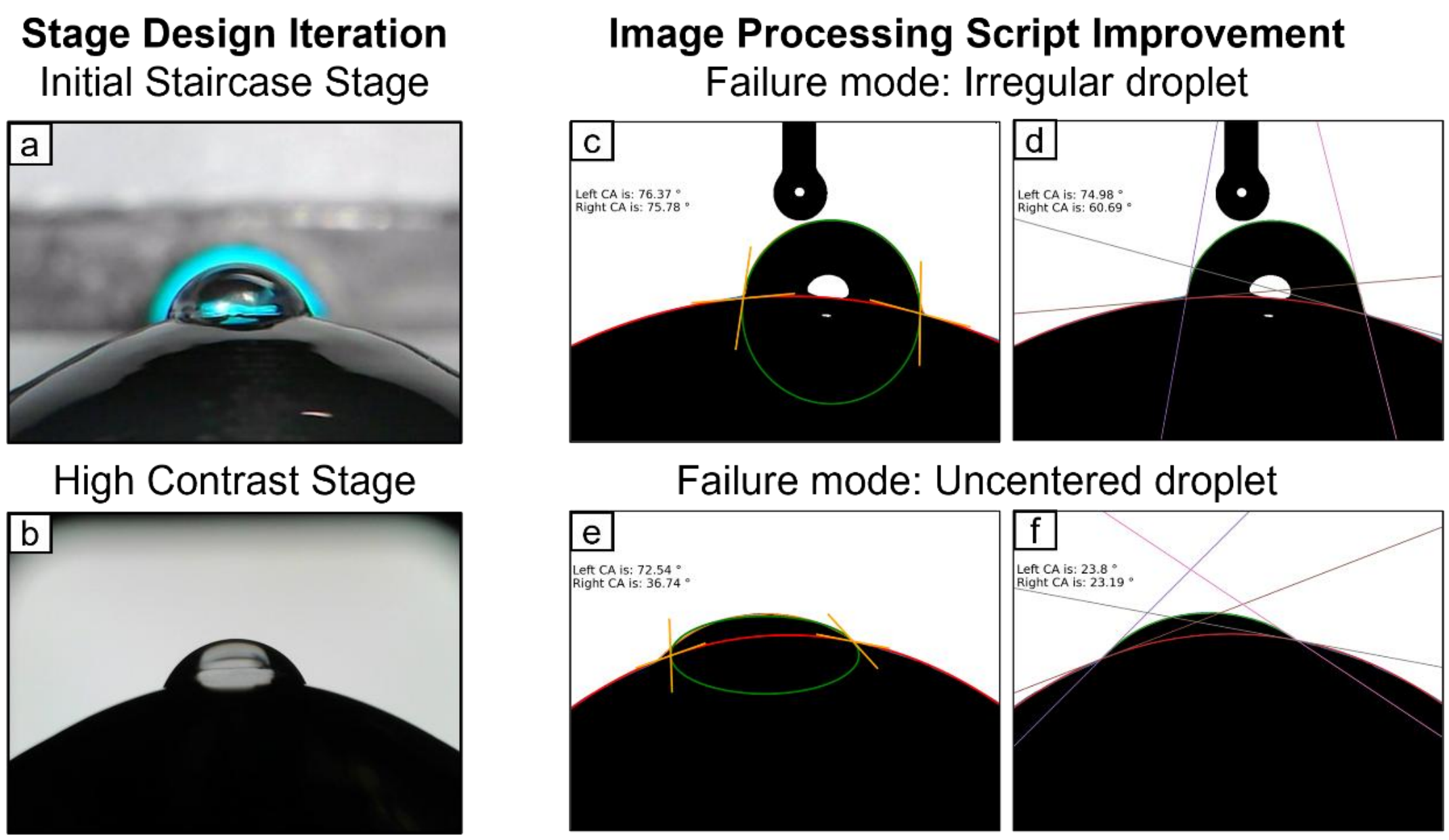

Figure 7. Resulting raw and processed images according to hardware and software adjustments during the development of RAISE-Vision: a) image of droplet on senofilcon A lens taken with the early iteration of the RAISE-Vision stage and mobile camera, b) Image of droplet on senofilcon A lens taken with optimized RAISE-Vision stage configuration and high contrast lighting, Irregular droplet image processing failure mode handled by the c) elliptical fit method and d) tangent method. Uncentered droplet image processing failure mode handled by the e) elliptical fit method, and d) tangent method.

Next, images captured on the original staircase stage design had much background noise, shown in Figure 7a, which proved to be difficult to handle during image processing. Further, the LED strip and camera's ring lighting would cause reflections on the contact lens and droplet. To better mimic the setup of standard goniometers, the high contrast lighting configuration of the optimized stage was tested, which significantly reduced noise in the captured image (Figure 7b).

At first, an open-source image processing script was written to automatically determine CA on curved surfaces the same way it is calculated with commercial software, via an elliptical fit method. Results obtained using the in-house script replicated analysis by the commercial software on the goniometer with an average absolute difference of 1.7°. Though the ellipse method is widely accepted and used [39], [46], [77], [82], its accuracy depends on the assumption that the droplet is adequately represented by an ellipse [67]. Symmetrical drops are ideal for image processing scripts, but are not always observed on flat surfaces, let alone curved surfaces [39]. Since the overall droplet contour influences the ellipse fit, local pinned geometry near the contact line may not be accounted for as seen on the right solid-liquid-air interface in Figure 7c. This inaccurate fitting is more likely for irregularly shaped high CA droplets as elliptical approximations may overestimate the local droplet curvature at the solid-liquid-air interface. Within the custom image processing script, other occasional failure modes such as uncentered low CA droplets were unable to be fit with an ellipse without rotation, shown in Figure 7e. These shifts in droplet symmetry relative to the vertical axis through the centre of the curved substrate have also previously been cited as a source of experimental variation during the determination of CA [67].

To account for these occasional errors of the custom elliptical method, the tangent method was employed as a second image processing algorithm to determine CA. As showcased in Figure 7df, the tangent algorithm is able to accurately represent irregular and off centered droplets. Instead of determining CA using the global droplet contour, the tangent method relies on the endpoints of the isolated droplet contour to identify the solid-liquid-air interface. The algorithm described in Section 2.3.4 verifies whether the selected point of intersection generates tangents that accurately describe the droplet. The discrepancy in determined CA according to image analysis method is consistent with previous reports demonstrating that choice of fitting modes affects CA output [83]. For example, Zhang et al. measured the CA of a 5 μL water droplet on a superhydrophobic surface using ellipse fitting, tangent method and Young-Laplace fitting to be 156.5°, 152.0° and 179.8° respectively [84]. The lower CAs obtained by the tangent method may be attributed to the discrete pixel representation of the droplet contour extracted via boundary tracing [52]. Because the droplets do not have a clear reflection on the contact lens surface, the exact location of the three-phase contact line may also be more difficult to detect as the visual contrast is reduced in the area [39]. Therefore, CAs directly calculated via tangents of the isolated droplet contour are more sensitive to image resolution, image contrast and camera focus than CAs calculated from the tangents of continuous elliptical curves [39], [83].

Because of the widespread and generally robust use of the elliptical fit method, it is suggested that the tangent method only be used in cases where standard ellipses do not properly represent the droplet shape. The integration of both methods in the presented combined image processing script improves the platform's tolerance to experimental variability. For non-ideal surfaces, local minimums in surface energy can cause the droplet to pin, leading to the potential of irregularly shaped droplets [78], [79]. Additionally, the application of droplets on curved surfaces may not always be centered, leading to potentially skewed droplets with inequal left and right CA [67], [85]. This combined approach reduces the dependence on perfect experimental execution and will improve the analysis of real, non-homogeneous surfaces.

Overall, the custom image processing script is able to account for a range of observed droplets, measuring high (> 90°) and low CA (> 20°) with high accuracy. As an automated tool, this script aims to enhance the reproducibility of measurements on curved surfaces, limits operator dependence and facilitates a smooth automated pipeline between data collection, analysis and storage. Further, the script's ability to automatically process images would be beneficial for other types of wettability analysis like dynamic CA where many frames must be processed during a single measurement [54].

## 4.2 Effect of Air Exposure on Senofilcon A Wettability

It is well established that surface properties of hydrogels depend on the orientation of hydrophobic and hydrophilic moieties at the surface, rather than the molecular configuration of the polymer, and is distinct from the bulk properties [51], [86], [87]. The senofilcon A material experienced about a 20° increase in CA over 10 sequential CA measurements (9 mins elapsed air exposure). This changing wetting behaviour can be explained by the segmental mobility of swollen SiHy systems and the reorientation of hydrophobic functional groups towards the lens surface when it is exposed to air or other hydrophobic environments [35], [87].

The initial cycles of the wetting profile exhibit the transient surface state of the SiHy materials as surface configuration changes to adjust from the initial equilibrium within a liquid medium to a new equilibrium with the surrounding air environment [86]. The lower CA observed during the first cycle in the wetting profile may also be attributed to surfactant residue from the blister pack solution reducing the surface tension of the applied droplet [51].

After the second droplet cycle, the CA of senofilcon A generally remained stable, indicating the material approaches surface state equilibrium with the surroundings quickly [88], which may be attributed to its internal wetting agent, polyvinyl pyrrolidone (PVP), spread through the lens matrix [89]. PVP is a highly hydrophilic polymer capable of forming many hydrogen bonds with water (25-35 hydrogen bonds per PVP unit) at its side chains [90], [91]. With moisture retained in the bulk of the lens, the hydrophilic moieties of the lens polymers would be encouraged to face towards the bulk, while the hydrophobic moieties would tend towards the hydrogel-air interface, causing relatively high CA compared to other lens materials [86]. Further, N,N-dimethylacrylamide (DMA), which is one of the main constituents of senofilcon A, is highly effective at increasing the EWC and lowering the modulus of contact lenses, but has been linked to poor surface wettability outcomes [92], [93], [94].

Further testing of other contact lenses would enable further dissection of the effect of varying material composition, wetting technologies and surface treatments on the contact lens wettability profile.

## 4.3 Performance of RAISE-Vision

The wettability characteristics of contact lenses are widely measured and studied in relation to their influence on tear spreading, and tear film stability [41], [51]. A major limitation of the sessile drop CA measurement method within the contact lens field is the lack of protocol standardization [41], which limits the comparability of wettability performance studies. The results from the benchmarking test indicate that the RAISE-Vision platform has been tuned to replicate measurements manually made by a skilled operator, while delivering added benefits of automation. Aside from offloading repetitive tasks from laboratory scientists, who can now divert their focus to more mentally stimulating tasks, such as observing live wetting behaviour, automating all workflow steps also removes the likelihood of inconsistencies or manual errors due to operator variability or fatigue [69], [95]. Further, because the RAISE-Vision workflow only requires operator involvement at the beginning and end of each run, one operator could realistically monitor multiple RAISE-Vision platforms running at the same time on different Opentrons OT-2. This could significantly increase throughput compared to the manual protocol.

Within RAISE-Vision, decision gates during the lens pre-processing and sessile drop measurement are removed, improving the repeatability of results. The automated CA determination algorithm also limits user-dependency in image processing, which has previously been cited as a source of poor reproducibility in CA measurement [51]. Beyond the consistency of automation, the initial drying protocol employed by the robot is advantageous as the tested surface at the apex of the contact lens is preserved. The programmed wicking action of the KimWipe inserts avoids uncontrolled over-drying, or fibre deposition commonly associated with manual sample preparation. Collectively, the features of RAISE-Vision build a standardized platform for CA measurement and analysis within the contact lens field that can facilitate more comparable wettability analysis across studies, such as those focused on the development of novel lens materials and wetting agents, and understanding contact lens interaction with tear constituents and fouling [12], [50], [57], [62].

Contact lens surface wettability is also often measured as a single CA with large variations within the literature. However, by measuring CA over air exposure, the RAISE-Vision assay can monitor the dynamic progression of surface wettability, including surface states that could result from different sample preparation and measurement protocols. For instance, the wettability profile for senofilcon A lenses measured by RAISE-Vision encompasses a substantial range of the single point CAs reported for senofilcon A [63], [64], indicating that literature CA values may be mapped to surface states sampled with RAISE-Vision.

Though there are a number of platforms with sophisticated methods of simulating blink cycles [28], [96], [97], even those which simulate a dynamic tear film layer [28], the RAISE-Vision assay is presented as an accelerated method of understanding contact lens wettability throughout air exposure without tear flow simulation. Eftimov et al. carried out a study to determine the effects of blink-like desiccation/rehydration cycles in phosphate-buffered saline (PBS) on the wettability of SiHy lenses over 16 hours [61]. Within their study, they observed that senofilcon A initially had a CA of 66.7° which increased to 83.3° by the 4$^{th}$ hour of cycling. This wetting state remained relatively stable for up to 10 hours of simulated wear, after which the CA slowly increased to 94.5° by the 16$^{th}$ hour of cycling. RAISE-Vision generated comparable observations to this extended study, as the wetting profile of senofilcon A lenses started with a CA of 62.6 ± 5.5° and increased to 83.4 ± 2.1° by the 10$^{th}$ cycle, just 9 minutes of air exposure after the initial measurement. By automating routine CA measurements, data management is also standardized, which can improve traceability of executed protocols, simplify downstream analysis, and facilitate smooth collaboration within research teams [71]. These results support the RAISE-Vision assay as an accelerated method to measure wettability change for preliminary investigation of contact lens material performance and development, while optimizing resource utilization.

### 4.4 Limitations and Future Directions

Currently, the RAISE-Vision platform is limited to static CA measurement, which may be difficult to reproduce due to droplet settling in local energy minima, or metastable wetting states on the contact lens surface [40]. Adding capabilities to measure the advancing and receding CA, which represent the highest and lowest CA in the hysteresis range, within RAISE-Vision would be an appropriate next step to provides more repeatable and complete characterization of the lens surface [40]. These measurements would be especially relevant to the tear wetting behaviour over a contact lens, as the receding CA have been thought to gauge the likeliness of tear film breakup during eye opening [45], while the advancing CA is an indicator of the ability of the tear film to spread over the lens during blinking [34], [45].

Within this study, only senofilcon A contact lenses were tested. However, there are many commercial contact lenses made of conventional hydrogel and silicone hydrogel materials that employ various wetting technologies, including plasma surface treatment, internal wetting agents and water gradient technology [17], [20], [89]. A broader range of contact lens types can be evaluated using the RAISE-Vision platform to understand differences in wetting behaviour with changes in material composition and surface properties.

Though measurements were made with Milli-Q water in this study, the platform is not limited to this liquid. Surface active biological components such as mucins, lysozyme, lipids and proteins can interact with the contact lens to affect wettability [47], [64]. Future work could investigate wettability measurements with artificial tear solution or phosphate buffered solution, which may be more physiologically relevant as they better mimic the osmolarity, pH and constituents of the tear film [61].

Currently, the automated CA measurements are taken in ambient conditions. However, changes in external factors, such as temperature and humidity, have been shown to affect measured CA on other hydrogel and non-ideal surfaces [98], [99]. Other contact lens properties, such as dimensions, modulus and dehydration are also sensitive to temperature and humidity differences [100], [101]. Future improvements to RAISE-Vision, may involve incorporating blink models, automating lens handling and implementing temperature and humidity controls, all of which are expected to further reduce measurement variability and enhance reliability and biorelevance of generated data.

## 5.0 Conclusion

Overall, RAISE-Vision, is the first automated sessile drop CA measurement platform tailored to contact lens wettability analysis over air exposure. Custom hardware components, specifically the pick-and-place drying apparatuses, were developed to allow for consistent and repeatable sample preparation and droplet removal. An open-source custom image processing script was developed to determine CA on a curved surface without operator input via the elliptical fit and tangent methods. The RAISE-Vision assay was used to capture the wetting profile of senofilcon A contact lenses over 10 sequential CA measurements. These results corresponded well to a similar manual workflow and other extended air exposure studies. Overall, by standardizing contact lens wettability analysis, this automated tool aims to support the understanding of contact lens surface wetting characteristics and aid early stage development of new materials and wetting technologies [5], [28], [102].